\documentclass[10pt,twocolumn,letterpaper]{article}

\usepackage{cvpr}
\usepackage{times}
\usepackage{epsfig}
\usepackage{graphicx}
\usepackage{amsmath}
\usepackage{amssymb}
\usepackage{color}
\usepackage{algorithmic}
\usepackage{algorithm}
\usepackage{enumitem}

\usepackage[pagebackref=true,breaklinks=true,letterpaper=true,colorlinks,bookmarks=false]{hyperref}

\def\bsx{$\sigma^B_{d}$}
\def\bsr{$\sigma^B_{rgb}$}
\def\fsx{$\sigma^F_{d}$}
\def\fsr{$\sigma^F_{rgb}$}

\newcommand{\argmax}{\operatornamewithlimits{argmax}}

\newcommand{\sumu}{\displaystyle\sum}

\newcommand{\e}{\emph}
\newcommand{\tb}{\textbf}
\newcommand{\tcb}{\textcolor{blue}}

\cvprfinalcopy % *** Uncomment this line for the final submission

\def\cvprPaperID{1916} % *** Enter the CVPR Paper ID here

\ifcvprfinal\fi

\begin{document}

%%%%%%%%% TITLE
\title{Background Modeling Using Adaptive Pixelwise Kernel Variances\\ in a Hybrid Feature Space}

\author{Manjunath Narayana\\
{\tt\small narayana@cs.umass.edu}
\and 
Allen Hanson\\
{\tt\small hanson@cs.umass.edu}\\
University of Massachusetts, Amherst\\
\and
Erik Learned-Miller\\
{\tt\small elm@cs.umass.edu}\\
%\and
%Allen Hanson\\
%University of Massachusetts, Amherst\\
%{\tt\small hanson@cs.umass.edu}
%\and
%Erik Learned-Miller\\
%University of Massachusetts, Amherst\\
%{\tt\small elm@cs.umass.edu}
}

\maketitle
\thispagestyle{empty}

%%%%%%%%% ABSTRACT
\begin{abstract}
Recent work on background subtraction has shown developments on two major
fronts. In one, there has been increasing sophistication of probabilistic
models, from mixtures of Gaussians at each pixel~\cite{Stauffer99}, to kernel
density estimates at each pixel~\cite{Elgammal00}, and more recently to joint
domain-range density estimates that incorporate spatial
information~\cite{Sheikh05}.  Another line of work has shown the benefits of
increasingly complex feature representations, including the use of texture
information, local binary patterns, and recently scale-invariant local ternary
patterns~\cite{Li10}. In this work, we use joint domain-range based
estimates for background and foreground scores and show that dynamically
choosing kernel variances in our kernel estimates at each individual pixel can
significantly improve results. We give a heuristic method for selectively
applying the adaptive kernel calculations which is nearly as accurate as the
full procedure but runs much faster.
We combine these modeling improvements with recently developed complex
features~\cite{Li10} and show significant improvements on a standard
backgrounding benchmark.
\end{abstract}

%%%%%%%%% BODY TEXT
\section{Introduction}\label{sec:intro} 
Background modeling is often an important step in detecting moving objects
in video sequences~\cite{Stauffer99,Li03,Elgammal00}. 
A common approach to background modeling is to define and learn a background
distribution over feature values at each pixel location and then classify each
image pixel as belonging to the background process or not.
The distributions at each pixel may be modeled in a parametric manner using a
mixture of Gaussians~\cite{Stauffer99} (MoG) or using non-parametric kernel
density estimation~\cite{Elgammal00} (KDE). More recently, models that allow a
pixel's spatial neighbors to influence its distribution have been developed by
joint domain-range density estimation~\cite{Sheikh05}. These models that allow
spatial influence from neighboring pixels have been shown to perform better than
earlier neighbor-independent models.

Sheikh and Shah~\cite{Sheikh05} also show that the use of an explicit
foreground model along with a background model can be useful. In a manner
similar to theirs, we use a kernel estimate to obtain the
background and foreground scores at each pixel location using data samples from
a spatial neighborhood around that location from previous frames. The background score is computed as a kernel estimate depending on the distance
in the joint domain-range space between the estimation point and the samples in the background model. 
A similar estimate is obtained for the foreground score. Each pixel is then
assigned a (soft) label based on the ratio of the background and foreground
scores.

The variance used in the estimation kernel reflects the spatial and
appearance uncertainties in the scene. On applying our method to a data set with wide
variations across the videos, we found that choosing suitable kernel variances
during the estimation process is very important. With various
experiments, we establish that the best kernel variance could vary for different
videos and more importantly, even within a single video, different regions in the
image should be treated with different variance values.  For example, in a scene
with a steady tree trunk and leaves that are waving in the wind, the trunk
region can be explained with a small amount of spatial variance. The leaf regions
may be better explained by a process with a large variance.  Interestingly, when
there is no wind, the leaf regions may also be explained with a low variance. The
optimal variance hence changes for each region in the video and also across
time. This phenomenon is captured reasonably in MoG~\cite{Stauffer99} by use of
different parameters for each pixel which adapt dynamically to the scene
statistics, but the pixel-wise model does not allow a pixel's neighbors to
affect its distribution. Sheikh and Shah~\cite{Sheikh05} address the phenomenon by updating the model with data samples from the most recent frame.
We show that using location-specific variances in addition to updating the model greatly improves
background modeling.  Our approach with pixel-wise variances, which we call
the variable kernel score (VKS) method results in significant improvement 
over uniform variance models and state of the art backgrounding systems.

The idea of using a pixel-wise variance for background modeling is not new. 
Although Sheikh and Shah~\cite{Sheikh05} use a uniform variance, 
they discuss the use of variances that change as a function of the
data samples or as a function of the point at which the estimation is made.
%(called \e{sample-point estimator} and \e{balloon estimator} in the KDE literature
%respectively~\cite{Jones90, Mittal04}).  
Variance selection for KDE is a well
studied problem~\cite{Turlach93} with common solutions including mean integrated square error (MISE), asymptotic MISE (AMISE),
and the leave-one-out-estimator based solutions.  In the background subtraction
context, there has been work on using a different covariance at each pixel~\cite{Mittal04, Tavakkoli09}.
While Mittal and Paragios~\cite{Mittal04} require that the
uncertainties in the feature values can be calculated in closed
form, Tavakkoli et al.~\cite{Tavakkoli09} learn the covariances for each pixel from a
training set of frames and keep the learned covariances fixed for the
entire classification phase. 
We use a maximum-likelihood approach to select the best variance at each pixel
location. For every frame of the video, at each pixel location, the best
variance is picked from a set of variance values by maximizing the likelihood of
the pixel's observation under different variances. This makes our method a
\e{balloon estimator}~\cite{Mittal04}. By explicitly selecting the best variance from a range of
variance values, we do not require the covariances to be calculable in
closed-form and also allow for more flexibility at the classification stage. 

Selecting the best of many kernel variances for each pixel means increased
computation. One possible trade-off between accuracy and speed can be achieved
by a caching scheme where the best kernel variances from the previous frame are
used to calculate the scores for the current frame pixels.  If the resulting
classification is overwhelmingly in favor of either label, there is no need to
perform a search for the best kernel variance for that pixel. The expensive
variance selection procedure can be applied only to pixels where there is some
contention between the two labels. We present a heuristic that achieves
significant reduction in computation compared to our full implementation while
maintaining the benefits of adaptive variance.

Development and improvement of the probabilistic models is one of the two main
themes in background modeling research in recent years. The other theme is the
development of complex features like local binary~\cite{Heikkila06} and ternary
patterns~\cite{Li10} that are more robust than color features for the task of
background modeling.  Scale-invariant local ternary patterns~\cite{Li10} (SILTP)
are recently developed features that have been shown to be very robust to
lighting changes and shadows in the scene. By combining color features with
SILTP features in our adaptive variance kernel model,
we bring together the best ideas from both themes in the field and
achieve state of the art results on a benchmark data set.

The main contributions of this paper are:
\begin{enumerate}[noitemsep, topsep=0pt, partopsep=0pt]
\item A practical scheme for pixel-wise variance selection for background
modeling.
\item A heuristic for selectively updating variances to improve speed further.
\item Incorporation of complex SILTP features into the joint domain-range kernel
framework to achieve state of the art results.
\end{enumerate}
%The main contributions of this paper are:\\ 
%(1) A practical scheme for pixel-wise variance selection for background
%modeling.\\
%(2) A heuristic for selectively updating variances to improve speed further.\\
%(3) Incorporation of complex SILTP features into the joint domain-range kernel
%framework to achieve state of the art results.
%(4) Extensive experiments on a standard benchmark showing both state of the art
%results and analyzing the effects of various aspects of our system.

The paper is organized as follows. Section \ref{sec:BG_FG} discusses our
background and foreground models. Dynamic adaptation of kernel variances is
discussed in Section \ref{sec:kernel}. Results and comparisons are
in Section \ref{sec:results}. An efficient algorithm is discussed in Section
\ref{sec:cache}. We end with a discussion in Section~\ref{sec:disc}.  
\section{Background and foreground models}\label{sec:BG_FG}
In a video captured by a static camera, the pixel values are influenced by the
background phenomenon, and new or existing foreground objects. We refer to any phenomenon
that can affect image pixel values as a process.
Like Sheikh and Shah~\cite{Sheikh05}, we model the background and foreground processes
using data samples from previous frames. The scores for the
background and foreground processes at each pixel location are calculated using
contributions from the data samples in each model. One major difference
between Sheikh and Shah and our model is that we allow ``soft~labeling'', i.e. the data samples contribute probabilistically to the background score depending on the samples'
probability of belonging to the background.

Let a pixel sample $a = [a_x a_y a_r a_g a_b]$, where $(a_x, a_y)$ are the
location of the pixel and $(a_r, a_g, a_b)$ are the red, green, and blue values
of the pixel.
In each frame of the video, we compute background and foreground scores using
pixel samples from the previous frames. The background model consists of the samples 
$B = \{b_i:i\!\in\![1\!\!:n_B]\}$ and foreground samples are
$F = \{f_i:i\!\in\![1\!:\!n_F]\}$, with $n_B$ and $n_F$ being the number of background
and foreground samples respectively, and $b_i$ and $f_i$ being pixel samples
obtained from previous frames in the video.
Under a KDE model~\cite{Sheikh05}, the likelihood of the sample under the background model is
\begin{equation}\label{eq:kde_bg}
P(a|bg;\sigma^B) = \frac{1}{n_B}\sum_{i=1:n_B} G(a-b_i; \sigma^B),
\end{equation}
where $G(x; \sigma)$ is a multivariate Gaussian with zero mean and
covariance $\sigma^B$.
\begin{equation}
G(x; \sigma) =
(2\pi)^\frac{-D}{2}|\sigma|^\frac{-1}{2}\exp(\frac{-1}{2}x^T\sigma^{-1}x),
\end{equation}
where $D$ is the dimensionality of the vector $x$.

In our model, we approximate the background score at sample $a$
as
\begin{equation}\label{eq:score_bg}
\begin{split}
&S_B(a; \sigma^B_{d}, \sigma^B_{rgb})\!=\!\frac{1}{N_B}\sumu_{i=1}^{n_B}\big\{G([a_r a_g a_b]\!-\\
&\![b_{ir} b_{ig} b_{ib}]; \sigma^B_{rgb})
\times G([a_x a_y]\!-\![b_{ix} b_{iy}]; \sigma^B_{d}) \times P(bg|b_i)\big\}.
\end{split}
\end{equation}
$N_B$ is the number of frames from which the background samples have been collected,
$\sigma^B_{d}$ and $\sigma^B_{rgb}$ are two and three dimensional background
covariance matrices in spatial and color dimensions respectively. A large
spatial covariance allows neighboring pixels to contribute more to the
score at a given pixel location. Color covariance allows for some color
appearance changes at a given pixel location. Use of $N_B$ in the denominator compensates for the different lengths of the background and foreground models. 

The above equation basically sums the contribution from each
background sample based on its distance in color space, weighted
by its distance in spatial dimensions and the probability of the
sample belonging to the background. 

The use of $P(bg|b_i)$ in Equation \ref{eq:score_bg} and normalization by the
number of frames as opposed to the number of samples means that the score does not
sum to $1$ over all possible values of $a$. Thus, the score, although similar to
the likelihood in Equation \ref{eq:kde_bg}, is not a probability distribution.
%However, it has the advantage that data samples can contribute probabilistically to
%the score of the background label depending on their probability of being from
%the background process.

A similar equation holds for the foreground score: 
\begin{equation}\label{eq:score_fg}
\begin{split}
&S_F(a; \sigma^F_{d}, \sigma^F_{rgb})\!=\!\frac{1}{N_F}\sumu_{i=1}^{n_F}\big\{G([a_r a_g a_b]\!-\\
&\![f_{ir} f_{ig} f_{ib}]; \sigma^F_{rgb}) \times G([a_x
 a_y]\!-\![f_{ix} f_{iy}]; \sigma^F_{d}) \times P(f\!g|f_i)\big\}.
\end{split}
\end{equation}
$N_F$ is the number of frames from which the foreground samples have been collected,
$\sigma^F_{d}$ and $\sigma^F_{rgb}$ are the covariances associated with the
foreground process.

However, for the foreground process, to account for emergence of new 
colors in the scene, we mix in a constant contribution independent of the
estimation point's and data samples'  color values. We assume that
each data sample in a pixel's spatial neighborhood contributes a constant
value $u$ to the foreground score.
%In terms of likelihood, this can
%be thought of as placing a uniform distribution at each location in the pixel's
%neighborhood. 
The constant contribution $U_F(a)$ is given by
\begin{equation}\label{eq:uniform_fg}
\begin{split}
U_F(a; \sigma^F_{d}) &= \!\frac{1}{N_F}\sumu_{i=1}^{n_F}\{u \times G([a_x
 a_y]\!-\![f_{1x} f_{1y}]; \sigma^F_{d})\}.
\end{split}
\end{equation}

%Note that $U_F$ is independent of the color observations of the estimation point
%$(a_r, a_g, a_b)$ and the data samples $(f_{ir}, f_{ig}, f_{ib})$. Thus it adds
%a uniform value to the foreground score of any observation. This is akin to
%mixing a uniform distribution in a KDE estimate.
%We use \footnote{Although each color can take $256$ values, we found that using
%$256^{-3}$ as the constant factor resulted in most foreground pixels
%going undetected. We justify using $100^{-3}$ by pointing out that the
%three color bands are not independent and that the color entropy in
%natural images has been found to be around $16$ bits~\cite{Charrier96}
%(much lower than the 24 bits that are used to represent color).}

We get a modified foreground score by including the constant contribution:
\begin{equation}\label{eq:score_fg2}
\begin{split}
\hat{S}_F&(a; \sigma^F_{d}, \sigma^F_{rgb}) =\\
         &\alpha_F\times U_F(a; \sigma^F_{d}) +
         (1-\alpha_F)\times S_F(a; \sigma^F_{d}, \sigma^F_{rgb}).
\end{split}
\end{equation}
$\alpha_F$ is a parameter that represents the amount of mixing between the constant 
contribution and the color dependent foreground score. $u$ is set to $10^{-6}$ and $\alpha$ is set to $0.5$ for our experiments.
%To reduce complexity, all the covariance matrices are assumed to be diagonal
%matrices and hence can be parameterized by just the spatial and color dimension
%variances. 

To classify a particular sample as background or foreground, we can use a Bayes-like formula:
\begin{equation}\label{eq:bayes_score}
\begin{split}
&P(bg|a) = \frac{S_B(a; \sigma^B_{d}, \sigma^B_{rgb})}{S_B(a; \sigma^B_{d},
        \sigma^B_{rgb}) + \hat{S}_F(a; \sigma^F_{d},\sigma^F_{rgb})}\\
&P(f\!g|a) = 1-P(bg|a).
\end{split}
\end{equation}

Adding the constant factor $U$ to the foreground score (and hence to the
denominator of the Bayes-like equation) has the interesting property
that when either one of the foreground or background scores is significantly
larger than $U$, $U$ has little effect on the classification. However, if both the background and
foreground scores are less than $U$, then Equation~\ref{eq:bayes_score} will
return a low value as $P(bg|a)$. Hence, an observation that
has very low background and foreground scores will be classified as
foreground. This is desirable because if a pixel observation is not well explained by either model, it is natural to assume that the pixel is a result of a new object in the scene and is hence foreground. In terms of likelihoods, adding the constant factor to the
foreground likelihood is akin to mixing it with a uniform distribution.

%\tb{Spatial neigborhood extent} - 
%Samples that are distant in spatial dimensions contribute very little in the
%score equations \ref{eq:score_bg} and \ref{eq:score_fg}. 
%%Thus, we can define the neighborhood for each pixel to be
%%such that pixels outside the neighborhood contibute almost zero value in the
%%equation \ref{eq:kde_bg}. 
%Since we use a Gaussian function to weight the spatial distance and 95\% of the
%mass of a Gaussian distribution falls within two standard deviations of the mean, any
%samples that are more than twice the standard deviation away from the sample
%$a$ in spatial dimension may be ignored. Thus, the choice of $n_B$ and $n_F$ are
%defined by the spatial variances.
\subsection{Model initialization and update}\label{subsec:bg_update}
To initialize the models, it is assumed that the first few frames (typically
$50$) are all background pixels. The background model is populated using
pixel samples from these frames. In order to improve efficiency, we sample $5$
frames at equal time intervals from these $50$ frames.
The foreground model is initialized to have no
samples. The modified foreground score (Equation \ref{eq:score_fg2}) enables colors that are
not well explained by the background model to be classified as foreground, thus
bootstrapping the foreground model. Once the pixel at location
$(a_x, a_y)$ from a new frame is classified using Equation \ref{eq:bayes_score},
the background and foreground models at the location $(a_x, a_y)$ can then
be updated with the new sample $a$. Background and foreground samples
at location $(a_x, a_y)$ from the oldest frame in the models are replaced by $a$.
Samples from the previous $5$ frames are maintained in memory as the foreground model samples. 
The label probabilities of the background/foreground from Equation \ref{eq:bayes_score}
are also saved along with the sample values for subsequent use in the Equations \ref{eq:score_bg} and \ref{eq:score_fg}. 

One consequence of the update procedure described above is that when a large foreground
object occludes a background pixel at $(a_x, a_y)$ for more than $50$ frames, 
all the background samples in the spatial neighborhood of $(a_x, a_y)$ 
are replaced by these foreground samples that have very low $P(bg|b_i)$ values. 
This causes the pixel at $(a_x, a_y)$ to be
misclassified as foreground even when the occluding foreground object has moved
away (because the background score will be extremely low due to the influence 
of $P(bg|b_i)$ in Equation \ref{eq:score_bg}). 
To avoid this problem, we replace the background sample from location
$(a_x,a_y)$ in the oldest frame in the background model with the new sample $a$
from the current frame only if $P(bg|a)$ estimated from Equation~\ref{eq:bayes_score} 
is greater than $0.5$.  

In our chosen evaluation data set, there are several videos with moving
objects in the first $50$ frames. The assumption that all these pixels are
background is not severely limiting even in these videos. The model update
procedure allows us to recover from any errors that are caused by the presence
of foreground objects in the initialization frames.
%The uniform distribution component in equation
%\ref{eq:kde_fg2} allows for detection of new colors that are not sufficiently
%explained by the bg model.

%The KDE implementation of \DF makes it very similar to KDE in~\cite{Sheikh05},
%with the exception of using different kernel variances (\bsx, \bsr) and (\fsx,
%\fsr) for the bg and fg models respectively.
%$(\Sigma_{B_{RGB}},\Sigma_{B_{XY}})$ and $(\Sigma_{F_{RGB}}, \Sigma_{F_{XY}})$ 
\subsection{Using MRF to clean the classification}\label{sec:MRF}
Similar to Sheikh and Shah~\cite{Sheikh05}, we use a Markov random
field (MRF) defined over the posterior label probabilities of the 4-neighbors of each
pixel and perform the min-cut procedure to post-process the labels. The
$\lambda$ interaction factor between the nodes was set to $1$ for all our
experiments. 

\section{Pixel-wise adaptive kernel variance selection}\label{sec:kernel}
%\subsection{Background and foreground kernels}
\tb{Background and foreground kernels.}
Sheikh and Shah use the same kernel parameters for background and
foreground models. Given the different nature of the two processes, it is
reasonable to use different kernel parameters. For instance, foreground objects
typically move between 5 and 10 pixels per frame in the I2R~\cite{Li03} data set, whereas
background pixels are either stationary or move very little.  Hence, it is useful
to have a larger spatial variance for the foreground model than for the
background model. 
%If it is known that the video is taken from a static camera, ideally we would
%need no variance in the spatial dimension for the background model. However, it
%is useful to have some small amount of spatial variance for the background
%model. This spatial variance allows for neighboring pixels to explain pixels
%that may have been exposed after long occlusion time.
%The choice of kernel variances is an important aspect in KDE background models. 

%\subsection{Optimal kernel variance for all videos}\label{subsec:opti_kern}
\tb{Optimal kernel variance for all videos.}
In the results section, we show that for a data set with large variations like
I2R~\cite{Li03}, a single value for kernel variance for all videos 
is not sufficient to capture the variability in all the
videos.  
%In the absence of sufficient ground truth data, choosing an
%optimal kernel variance for each video is a difficult task.  

\begin{figure}[t]
\begin{center}
   \includegraphics[width=1.0\linewidth]{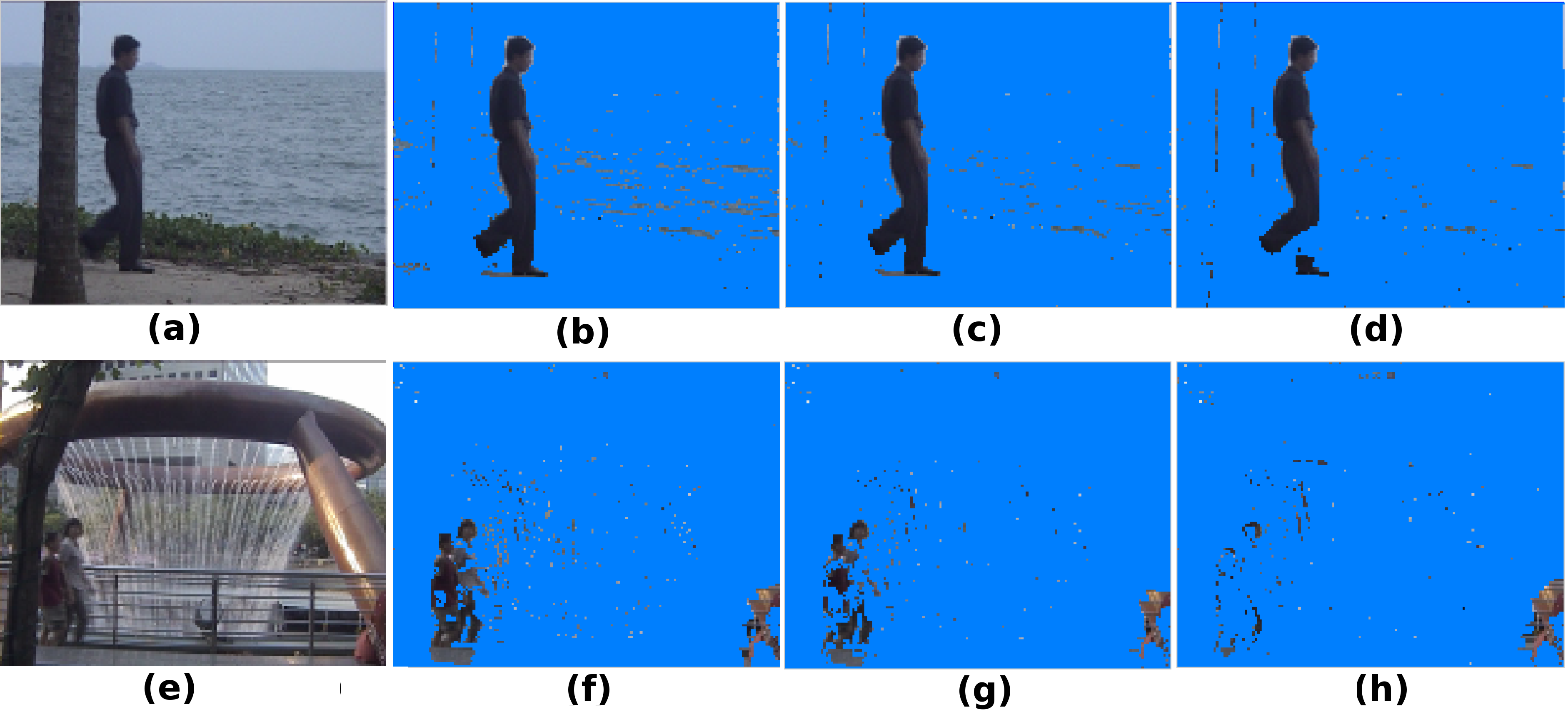}
\end{center}
   \caption{Two video sequences classified using increasing values of spatial kernel variance. 
       \tb{Column 1:} Original image. \tb{Column 2:} \bsx$=1/4$. \tb{Column 3:} \bsx$=3/4$. \tb{Column 4:} \bsx$=12/4$.}
\label{fig:sigma_ex}
\end{figure}

%\subsection{Uniform kernel variance for single video}\label{subsec:uni_kern} 
\tb{Variable kernel variance for a single video.}
As explained in the introduction, different parts of the scene may have
different statistics and hence need different kernel variance values. For
example, in Figure \ref{fig:sigma_ex}a to \ref{fig:sigma_ex}d, having a high
spatial dimension kernel variance helps in accurate classification of the
water surface pixels, but doing so causes some pixels on the person's leg to
become part of the background. Ideally, we would have different kernel variances
for the water surface pixels and the rest of the pixels. Similarly in the
second video (Figure \ref{fig:sigma_ex}e to \ref{fig:sigma_ex}h), having a high
kernel variance allows accurate classification of some of the fountain pixels as
background at the cost of misclassifying many foreground pixels. The figure also
shows that while the medium kernel variance may be the best choice for the first
video, the low kernel variance may be best for the second video.
%\subsection{Sharpening match principle in Distribution Fields}\label{subsec:sharpen} 
%\subsection{Optimal kernel variance for classification}\label{subsec:sharpen} 

\begin{figure}[t]
\begin{center}
   \includegraphics[height=0.5\linewidth]{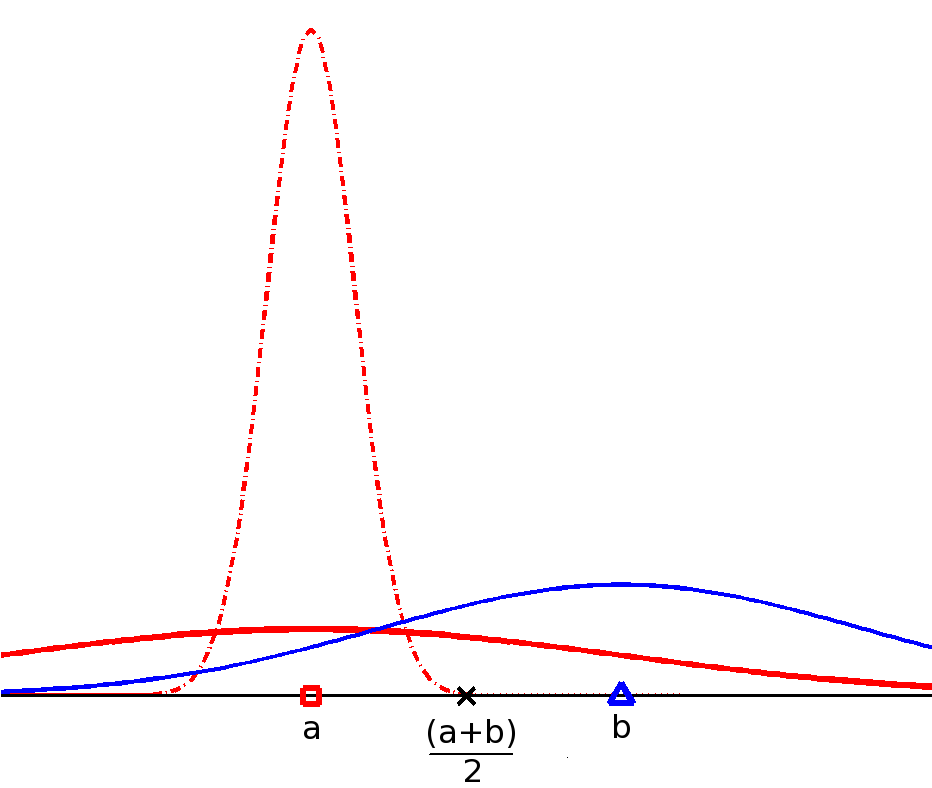}
\end{center}
   \caption{1-dimensional example shows the effect of the kernel variance in classification. Using a higher or lower variance at point `a' compared to point `b' can cause misclassification of the intermediate point between them.}
\label{fig:1D_variance_ex}
\end{figure}

\tb{Optimal kernel variance for classification.}
%The covariance matrix in color and spatial dimensions enable explaining
%uncertainties in color and position of the pixels in the scene. 
Having different variances for the background and foreground models reflects the
differences between the expected uncertainty in the two processes. However,
having different variances for the two processes could cause
erroneous classification of pixels. Figure \ref{fig:1D_variance_ex}
shows a 1-dimensional example where using a very wide kernel (high
variance) or very narrow kernel for the background process
causes misclassification. Assuming that the red point (square) is a
background sample and the blue point (triangle) is a foreground sample,
having a very low variance kernel (dashed red line) or a very high
variance (solid red line) for the background process makes the
background likelihood of the center point `x' lower than the
foreground likelihood. Thus, it is important to pick the optimal
kernel variance for each process during classification.
%This issue is also observed in tracking and using a "sharpening"
%match, where the kernel variances are reduced until maximum likelihood is
%achieved, alleviates the problem.\\ 

In order to address all four issues discussed above, we propose the use of
location-specific variances. For each location in the image, a range of kernel
variances is tried and the variance which results in the highest score is chosen
for the background and the foreground models separately.

The background score with location-dependent variances is
\begin{equation}\label{eq:var_score_bg}
\begin{split}
S_B(&a; \sigma^B_{d,a_x, a_y}, \sigma^B_{rgb, a_x,a_y}) = \\
    \frac{1}{N_B}&\sumu_{i=1}^{n_B}\{G([a_r a_g a_b]-[b_{ir} b_{ig} b_{ib}]; \sigma^B_{rgb,a_x,a_y})\\ 
    &\ \ \ \times G([a_x a_y]\!-\![b_{1x} b_{1y}]; \sigma^B_{d,a_x,a_y}) \times
        P(bg|b_i)\},
\end{split}
\end{equation}
where $\sigma^B_{d,x,y}$ and $\sigma^B_{rgb,x,y}$ represent the
location-specific spatial and color dimension variances at location $(x, y)$.

For each pixel location $(a_x, a_y)$, the optimal variance for the background
process is selected by maximizing the score of the background label at sample
$a$ under different variance values:
\begin{equation}\label{eq:max_bg_var}
\begin{split}
\{&\sigma^{B*}_{d,a_x,a_y}, \sigma^{B*}_{rgb,a_x,a_y}\} =\\
&\argmax_{\sigma^B_{d,a_x,a_y},\sigma^B_{rgb,a_x,a_y}}\!\!\!\!\!\! S_B(a;
        \sigma^B_{d,a_x,
        a_y}, \sigma^B_{rgb, a_x, a_y}).
\end{split}
\end{equation}
%Thus, for each pixel location $(x,y)$, for the bg model, the optimal variances
%$\sigma^*_{B_{XY}}(x,y)$ and $\sigma^*_{B_{RGB}}(x,y)$ are selected from a set
%of values $S_{B_{XY}}$ and $S_{B_{RGB}}$. 
Here, $\sigma^{B}_{d,a_x,a_y} \in R^B_{d}$ and $\sigma^{B}_{rgb,a_x,a_y} \in
R^B_{rgb}$. $R^B_{d}$ and $R^B_{rgb}$ represent the set of spatial and color dimension variances from which to choose the optimal variance.

%Similarly, for the fg variances, $\sigma^*_{F_{XY}}(x,y)$ and
%$\sigma^*_{F_{RGB}}(x,y)$ are selected from the sets $S_{F_{XY}}$ and
%$S_{F_{RGB}}$. The optimal variance calculation is performed for each pixel
%location for each frame in the video.
A similar procedure may be followed for the foreground score.
However, in practice, it was found that the variance selection procedure yielded large
improvements when applied to the background model and little improvement in the
foreground model.  Hence, our final implementation uses an adaptive kernel variance
procedure for the background model and a fixed kernel variance for the
foreground model.
\section{Results}\label{sec:results}
For comparisons, we use the I2R data set~\cite{Li03} which consists of $9$ videos
taken using a static camera in various environments. The data set offers various
challenges including dynamic background like trees and waves, gradual and sudden
illumination changes, and the presence of multiple moving objects. 
%Our method is
%not designed to handle sudden illumination change like turning a light switch
%on or off. The Lobby video in the data set is an example of such a scenario.
%Hence we omit this video from our analysis. 
Ground truth for $20$ frames in each video is provided with the data set. The
F-measure is used to measure accuracy~\cite{Li10}.
%\begin{equation}
%%F = \frac{2.recall.precision}{recall+precision} = \frac{2TP}{2TP+FN+FP}\\
%F = \frac{2\times recall\times precision}{recall+precision}\\
%\end{equation}

The effect of choosing various kernel widths for the background and foreground
models is shown in Table \ref{tbl:I2R_sigmas}. The table shows the F-measure for
each of the videos in the data set for various choices of the kernel variances.
The first 5 columns correspond to using a constant variance for each process at
all pixel locations in the video. Having identical kernel variances for the
background and foreground models (columns 1, 2) is not as effective as having
different variances (all other columns). Comparing columns 2 and 3 shows that
using a larger spatial variance for the foreground model than for the background
model is beneficial. Changing the spatial variance from 3 (column 3) to 1
(column 4) helps the overall accuracy in one video (Fountain). 
%The table reports overall accuracy for the video and
%does not show the difference in individual frames and pixels. 
%When individual
%frames are observed, variance settings from columns 1,2, and 4 often outperform
%the settings in column 3.  
Using a selection procedure where the best kernel
variance is chosen from a set of values gives the best results for most
videos (column 6) and frames.

Comparison of our selection procedure to a baseline method of using a standard
algorithm for variance selection in KDE (AMISE criterion\footnote{We use the
publicly available implementation from
http://www.ics.uci.edu/~ihler/code/kde.html.}) shows that the
standard algorithm is not as accurate as our method (column 7).
%This unusual result does not reflect the soundness of our method, but is mainly
%due to the fact that the standard method tends to learn the best variance
%from an unrestricted set of values and hence "explains" most pixels as
%belonging to background. 
Our choice for the variance values for spatial
dimension reflects no motion (\bsx$=1/4$) and very little motion (\bsx$=3/4$) for
the background, and moderate amount of motion (\fsx$=12/4$) for the foreground. For
the color dimension, the choice is between little variation (\bsr$=5/4$),
moderate variation (\bsr$=15/4$), and high variation (\bsr$=45/4$) for the
background, and moderate variation (\fsr$=15/4$) for the foreground. These choices
are based on our intuition about the processes involved.
%naked-eye observations from the data and hence better
%suited for the task of background subtraction in this data set. 
%The baseline method may well be more successful if this information is not known
%beforehand. The baseline method however takes 6 times as long to execute.
For videos that differ significantly from the videos we use, it is possible that the baseline AMISE method would perform better.

We would like to point out that ideally the variance value sets should be learned
automatically from a separate training data set. In absence of suitable training
data for these videos in particular and for background subtraction research in
general, we resort to manually choosing these values. This also
appears to be the common practice among researchers in this area.

%\onecolumn
\begin{table*}
\begin{center}
\begin{tabular}{|l|c|c|c|c|c|c|c|}
\hline
Column num               &(1)&(2)&(3)&(4)&(5)&(6)       &(7)\\
4*\bsx $\longrightarrow$ &3  &3  &3  &1  &3  &[1 3]     &AMISE\\
4*\bsr $\longrightarrow$ &15 &45 &45 &45 &15 &[5 15 45] &AMISE\\
4*\fsx $\longrightarrow$ &3  &3  &12 &12 &12 &[12]      &[12]\\
4*\fsr $\longrightarrow$ &15 &45 &45 &45 &15 &[15]      &[15]\\ 
\hline\hline
AirportHall     &$40.72$ &$59.53$ &$67.07$ &$63.53$ &$47.21$ &$\tb{70.44}$ &$53.01$\\
Bootstrap       &$49.01$ &$57.90$ &$63.04$ &$58.39$ &$51.49$ &$\tb{71.25}$ &$63.38$\\
Curtain         &$66.26$ &$83.33$ &$91.91$ &$89.52$ &$81.54$ &$\tb{94.11}$ &$52.00$\\
Escalator       &$20.92$ &$30.24$ &$34.69$ &$28.58$ &$22.65$ &$\tb{48.61}$ &$32.02$\\
Fountain        &$41.87$ &$51.89$ &$73.24$ &$74.58$ &$67.60$ &$\tb{75.84}$ &$28.50$\\
ShoppingMall    &$55.19$ &$60.17$ &$64.95$ &$62.18$ &$63.85$ &$\tb{76.48}$ &$70.14$\\
Lobby           &$22.18$ &$23.81$ &$25.79$ &$25.69$ &$25.06$ &$18.00$ &$\tb{36.77}$\\
Trees           &$30.14$ &$58.41$ &$73.53$ &$47.03$ &$67.80$ &$\tb{82.09}$ &$64.30$\\
WaterSurface    &$85.82$ &$94.04$ &$\tb{94.93}$ &$92.91$ &$94.64$ &$94.83$ &$30.29$\\
\hline\hline
Average         &$45.79$ &$57.70$ &$65.46$ &$60.27$ &$52.98$ & $\tb{70.18}$ & $47.82$\\
%Average without Lobby    & $28.39$ & $53.18$ &$52.28$ &$57.75$ &$63.70$\\
\hline
\end{tabular}
\end{center}
\caption{\e{F-measure} for different kernel variances. Using our selection
    procedure ( Column 6) results in the highest accuracy.}
\label{tbl:I2R_sigmas}
\end{table*}
%\twocolumn
Benchmark comparisons are provided for selected existing methods -
MOG~\cite{Stauffer99}, the complex foreground model~\cite{Li03} (ACMMM03),
and SILTP~\cite{Li10}.
%Although SILTP is originally
%introduced in~\cite{Li10}, the table results are from the SILS
%paper~\cite{Yuk11} who reported better performance. 
%Scale Invariant Local Ternary Patterns (SILTP) and Scale Invariant Local States
%(SILS) are ternary pattern-based features that are reported to be very robust
%to lighting changes and shadows.  
%Setting the kernel variances to constant values in our method corresponds to
%the KDE in~\cite{Sheikh05}. 
To evaluate our results, the posterior probability of the background label is
thresholded at a value of $0.5$ to get the foreground pixels.  Following the same
procedure as Liao et al.~\cite{Li10}, any foreground 4-connected components smaller than a
size threshold of $15$ pixels are ignored.

\begin{figure}
\begin{center}
   \includegraphics[width=1.0\linewidth]{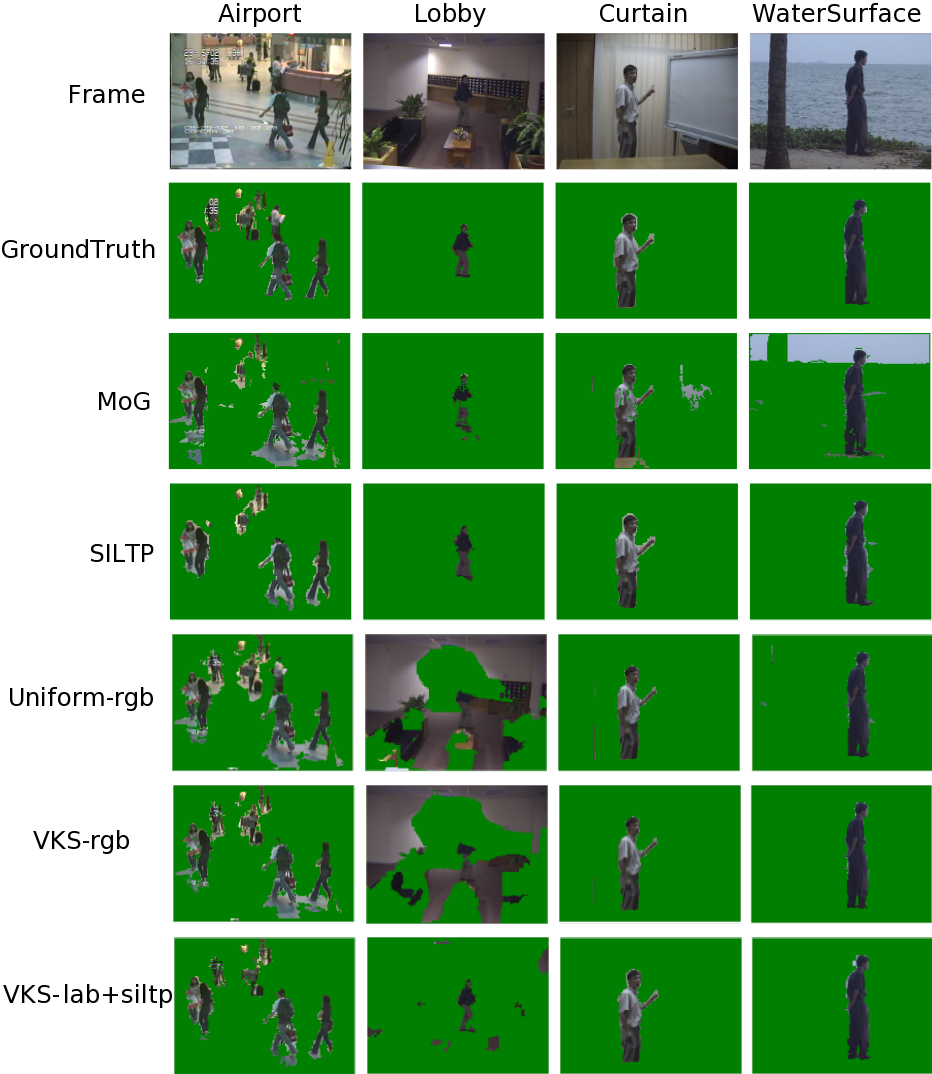}
\end{center}
%\label{fig:short}
\caption{Qualitative comparison of algorithms on image results reported in Liao et al.~\cite{Li10}.}
\label{fig:I2R_compare}
\end{figure}

Figure \ref{fig:I2R_compare} shows qualitative results for the same frames that
were reported by Liao et al.~\cite{Li10}. We present results for our kernel method with
uniform variances and adaptive variances with RGB features (Uniform-rgb and
VKS-rgb respectively), and adaptive variances with a hybrid feature space of LAB
color and SILTP features (VKS-lab+siltp).
Except for the Lobby video, the VKS results are better than other methods.  The
Lobby video is an instance where there is a sudden change in illumination in the
scene (turning a light switch on and off).  Due to use of an explicit foreground
model, our kernel methods misclassify most of the pixels as foreground and take
a long time to recover from this error. A possible solution for this case
is presented later. Compared to the uniform variance kernel estimates, we see
that VKS-rgb has fewer false positive foreground pixels. 

%This can be explained by the bg model adapting its
%kernel variance to explain as much of the background as possible. 
Quantitative results in Table \ref{tbl:I2R_compare} compare the F-measure scores
for our method against MoG, ACMMM03, and SILTP results as reported by Liao et al.~\cite{Li10}.
The table shows that methods that share spatial information (uniform kernel and
VKS) with RGB features give significantly better results than methods that use
RGB features without spatial sharing.
Comparing the variable kernel method to a uniform kernel method in the same
feature space (RGB), we see a significant improvement in performance for most
videos.
%F-measure scores show that our system performs significantly better than MoG and ACMMM03 on most videos.
%The qualitative results show that the performance of DF is close to the SILTP
%and
%SILS performance than  \ref{tbl:I2R_compare} suggest. The performance of SILTP
%vis-a-vis the DF is further discussed in the next section.
%\subsection{Discussion on SILTP}\label{subsec:SILTP}
Scale-invariant local ternary pattern (SILTP)~\cite{Li10} is a recent texture
feature that is robust to soft shadows and lighting changes. We believe SILTP
represents the state of the art in background modeling and hence compare our
results to this method. Scale-invariant local states~\cite{Yuk11} is a
slight variation in the representation of the SILTP feature. For comparison, we
use SILTP results from Liao et al. because in Yuk and Wong~\cite{Yuk11}, human 
judgement\footnote{This was learned via personal communication with the authors.} 
was used to vary a size threshold parameter for each video.  
We believe results from the latter fall under a different category of
human-assisted backgrounding and hence do not compare to our method where no
video-specific hand-tuning of parameters was done. 
Table \ref{tbl:I2R_compare} shows that SILTP is very robust to lighting changes
and works well across the entire data set. Blue entries in Table~\ref{tbl:I2R_compare} correspond
to videos where our method performs better than SILTP.
VKS with RGB features (VKS-rgb) performs well in videos that have few shadows
and lighting changes. 
%This difference in performance is mainly due to shadow regions being
%misclassified by use of RGB features. 
%RGB values are not known to be robust to shadows. 
%It is interesting to note that in the 4 videos in the
%data set where there is no significant lighting changes or shadows (Curtain,
%Fountain, Campus, WaterSurface), VKS-rgb significantly improves over
%SILTP in 3 videos.  Also, SILTP is known to produce
%"holes"~\cite{Yuk11} in objects that are large and of homogenous texture. 
%This can be observed in the first column of figure \ref{fig:I2R_compare}. 
%In this data set, there are 3 videos with consistently large objects (Curtain,
%Campus, and WaterSurface).  Table \ref{tbl:I2R_compare} shows that VKS-rgb
%performs better than SILTP on all 3 videos.
Use of color features that are more robust to illumination change, like LAB
features in place of RGB helps in successful classification of the shadow regions
as background. Texture features are robust to lighting changes but not effective
on large texture-less objects. Color features are effective on large objects,
but not very robust to varying illumination. By combining texture features
with LAB color features, we expect to benefit from the strengths of
both feature spaces. Such a combination has proved useful in earlier
work~\cite{Yao07}. Augmenting the LAB features with SILTP features (computed at 3
resolutions) in the VKS framework (VKS-lab+siltp) results in an improvement in 
7 out of 9 videos (last column). 
The variance values used in our implementation are given in Table~\ref{tbl:VKS_parameters}.

\begin{table*}
\begin{center}
\begin{tabular}{|l|c|c|c|c|c|c|c|c|c|c|}
\hline
               & $4\!*\!\sigma^B_{d}$ & $4\!*\!\sigma^F_{d}$ & $4\!*\!\sigma^B_{rgb}$ & $4\!*\!\sigma^F_{rgb}$ & $4\!*\!\sigma^B_{l}$ & $4\!*\!\sigma^F_{l}$ & $4\!*\!\sigma^B_{ab}$ & $4\!*\!\sigma^F_{ab}$ & $4\!*\!\sigma^B_{siltp}$ & $4\!*\!\sigma^F_{siltp}$   \\
\hline
VKS-rgb       &[1,3] & 12  &[5, 15, 45] & 15 & -           & -   & -     & - & - & -\\
VKS-lab+siltp &[1,3] & 12  &     -      & -  & [5, 10, 20] & 15  & [4,6] & 4 & 3 & 3\\
\hline
\end{tabular}
\end{center}
\caption{Parameter values for VKS implementation.}
\label{tbl:VKS_parameters}
\end{table*}
%The variance values used for VKS were:\\ 
%($[\ \ ]$ indicates selection set)\\
%For both models:
% $4\!*\!\sigma^B_{d}=[1, 3], 4\!*\!\sigma^F_{d}=12$\\
%rgb model : $4\!*\!\sigma^B_{rgb}=[5, 15, 45], 4\!*\!\sigma^F_{rgb}=15$\\
%%LAB model : $4\!*\!\sigma^B_{l}=[5, 10, 20], 4\!*\!\sigma^B_{ab}=[4, 6]\\
%%4\!*\!\sigma^F_{l}=15, 4\!*\!\sigma^F_{ab}=4$\\
%lab+siltp model : \\
%$4\!*\!\sigma^B_{l}=[5, 10, 20], 4\!*\!\sigma^B_{ab}=[4, 6],
%4\!*\!\sigma^B_{siltp}=3,\\
%4\!*\!\sigma^F_{l}=15, 4\!*\!\sigma^F_{ab}=4, 4\!*\!\sigma^F_{siltp}=3$\\
%For SILTP space, the XOR operation between the binary representations of two
%SILTP features gives the distance between them~\cite{Li10}.

%%It would be interesting to study why the combination of color and SILTP does not
%%outperform both color and SILTP on all videos. It must be noted that the SILTP
%%feature used in this paper is our own implementation and not the original code
%%from~\cite{Li10}. Results of our SILTP feature and algorithm implementation as
%%described in~\cite{Li10} are given in the last column in table
%%\ref{tbl:I2R_compare}. Comparison to these F-measure numbers shows that DF
%%method combining LAB and SILTP indeed outperforms both LAB and SILTP
%%individually in 7 out of 8 videos. Perhaps the discrepancy in column 8 is due to
%%differences in our implementation and the original implementation of SILTP.

\begin{figure}
\begin{center}
   \includegraphics[width=1.0\linewidth]{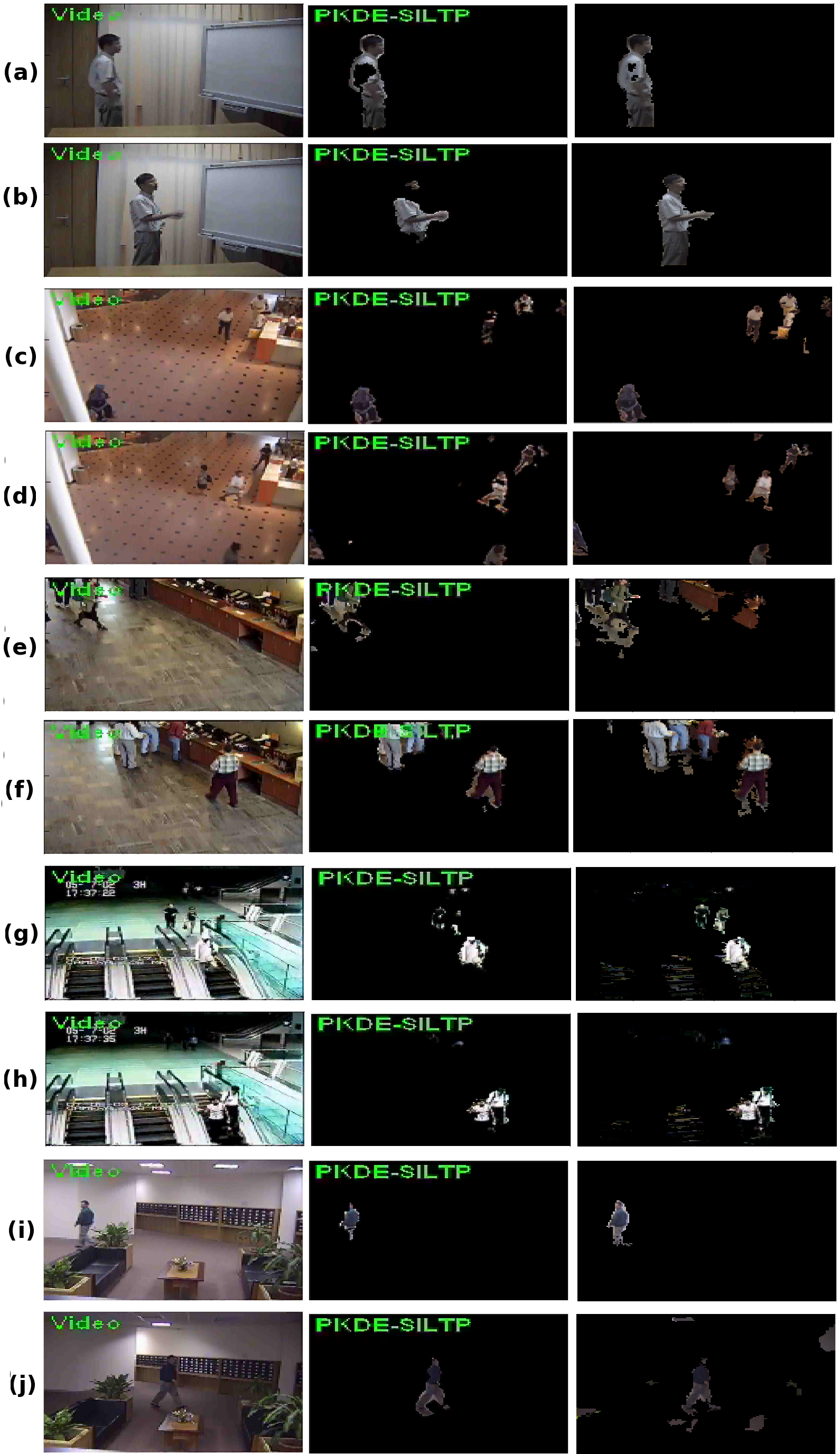}
\end{center} 
\caption{Comparison of VKS and SILTP results. 
    \tb{Column 1:} Original video. \tb{Column 2:} SILTP~\cite{Li10}. \tb{Column 3:} VKS-lab+siltp.}
\label{fig:DF_SILTP_compare}
\end{figure}

We also compare our results (VKS-lab+siltp) to the 5 videos that were submitted
as supplementary material by Liao et al.~\cite{Li10}. Figure
\ref{fig:DF_SILTP_compare} highlights some key frames that highlight the
strengths and weaknesses of our system versus the SILTP results. The common
problems with our algorithm are shadows being classified as foreground (row e) 
and initialization errors
(row e shows a scene where the desk was occluded by people when the background
 model was initialized. Due to the explicit foreground model, VKS takes some
 time to recover from the erroneous initialization). A common drawback with
SILTP is that large texture-less objects have ``holes'' in them (row a). Use of
color features helps avoid these errors.  The SILTP system also loses
objects that stop moving (rows b, c, d, f). Due to the explicit modeling of the
foreground, VKS is able to detect objects that stop moving.

The two videos in the data set where our algorithm performs worse than SILTP are
the Escalator video (rows g, h) and the Lobby video (rows i, j). In the
Escalator video, our algorithm fails at the escalator steps due to large
variation in color in the region.

In the Lobby video, at the time of sudden illumination change, many pixels in
the image get classified as foreground. Due to the foreground model, these
pixels continue to be misclassified for a long duration (row j). The problem
is more serious for RGB features (Figure \ref{fig:I2R_compare} column 2). One
method to address the situation is to observe the illumination change from one
frame to the next. If more than half the pixels in the image change in
illumination by a threshold value of $T_I$ or more, we throw away all the
background samples at that instance and begin learning a new model from the
subsequent $50$ frames. This method allows us to address the poor performance
in the Lobby video with resulting F-measure values of $86.77$ for uniform-rgb,
$78.46$ for VKS-rgb, and $77.76$ for VKS-lab+siltp.  $T_I$ of $10$ and $2.5$
were used for RGB and LAB spaces respectively. The illumination change
procedure does not affect the performance of VKS on any other video in the
data set.
%Table showing our implementation of SILTP???.\\
%\onecolumn
\begin{table*}
\begin{center}
\begin{tabular}{|l|c|c|c|c|c|c|c|c|}
\hline
Video & ACMMM03 & MoG & SILTP~\cite{Li10} & uniform & VKS  & VKS       \\ 
      &         &     &                   & rgb     & rgb  & lab+siltp \\ 
\hline\hline
AirportHall     & $50.18$ & $57.86$ & $68.02$ & $67.07$ & $\tcb{70.44}$ & $\tcb{\tb{71.28}}$\\
Bootstrap       & $60.46$ & $54.07$ & $72.90$ & $63.04$ & $71.25$ &
$\tcb{\tb{76.89}}$\\
Curtain         & $56.08$ & $50.53$ & $92.40$ & $91.91$ & $\tcb{\tb{94.11}}$ & \tcb{$94.07$}\\
Escalator       & $32.95$ & $36.64$ & $\tb{68.66}$ & $34.69$ & $48.61$ & $49.43$\\
Fountain        & $56.49$ & $77.85$ & $85.04$ & $73.24$ & $75.84$ & $\tcb{\tb{85.97}}$\\
ShoppingMall    & $67.84$ & $66.95$ & $79.65$ & $64.95$ & $76.48$ & $\tcb{\tb{83.03}}$\\
Lobby           & $20.35$ & $68.42$ & $\tb{79.21}$ & $25.79$ & $18.00$ & $60.82$\\
Trees           & $\tcb{75.40}$ & $55.37$ & $67.83$ & \tcb{$73.53$} & $\tcb{82.09}$ &
$\tcb{\tb{87.85}}$\\
WaterSurface    & $63.66$ & $63.52$ & $83.15$ & $\tcb{\tb{94.93}}$ & \tcb{$94.83$} & \tcb{$92.61$}\\
%\hline\hline
%Average         & $57.84$ & $63.11$ & $XX.XX$ & $73.16$ & $76.70$ &$76.40$ & $80.14$ &$74.85$ &$$\\
%Average without Lobby    & $28.39$ & $53.18$ & $52.28$ & $57.75$ & $XX.XX$ & %$63.70$ &$XX.XX$ & $XX.XX$ &$$\\
\hline
\end{tabular}
\end{center}
\caption{\e{F-measure} on I2R data. VKS significantly outperforms other
    color feature-based methods and improves on SILTP texture features on most
    videos. Blue color indicates performance better than SILTP.}
\label{tbl:I2R_compare}
\end{table*}
\section{Caching optimal kernel variances from previous frame}\label{sec:cache}
A major drawback with trying multiple variance values at each pixel to select
the best variance is that the amount of computation per pixel increases
significantly. In order to reduce the complexity the algorithm, we use a scheme
where the current frame's optimal variance values for each pixel location for
both the background and foreground processes is stored
($\sigma^{*B_{cache}}_{x,y}, \sigma^{*F_{cache}}_{x,y})$
for each location $(x,y)$ in the image.
When classifying pixels in the next frame, these cached variance values are
first tried. If the resulting scores are very far apart, then it
is very likely that the pixel has not changed its label from the previous
frame. The expensive variance selection procedure is performed only at pixels
where the resulting scores are close to each other. 
%Algorithm \ref{alg:eff_var} explains the process.
%\begin{algorithmic}
%\FOR{ each pixel sample  $a = (a_x,a_y,a_r,a_g,a_b)$ in the current frame}
%\IF{$\frac{S_B(a; \sigma^{*B{cache}}_{a_x,a_y})}{S_B(a;
%        \sigma^{*B{cache}}_{a_x,a_y})} >\tau_{BF}$} 
%or $\frac{P(f\!g|a, \Sigma_{F_{cached}}(a_x, a_y))}{P(bg|a,
%\Sigma_{B_{cached}}(a_x, a_y))}>\tau_{FB}$,\\
%\STATE Find the label scores resulting from use of the cached variance
%values
%\ELSE
%\STATE Search over the values in the variance sets to pick the optimal variances\\
%\STATE Find the label scores using the optimal variances
%\ENDIF
%\ENDFOR
%\end{algorithmic}
\begin{algorithm}
\begin{algorithmic}
\FOR{ each pixel sample  $a = (a_x,a_y,a_r,a_g,a_b)$ in the current frame}
\IF{$\frac{S_B(a; \sigma^{*B{cache}}_{a_x,a_y})}{\hat{S_F}(a;
        \sigma^{*F{cache}}_{a_x,a_y})} >\tau_{BF}$} 
%or $\frac{P(f\!g|a, \Sigma_{F_{cached}}(a_x, a_y))}{P(bg|a,
%\Sigma_{B_{cached}}(a_x, a_y))}>\tau_{FB}$,\\
\STATE Compute the label scores resulting from use of the cached variance
values.
\ELSE
\STATE Search over the values in the variance sets to pick the optimal variances.\\
\STATE Compute the label scores using the optimal variances.
\ENDIF
\ENDFOR
\end{algorithmic}
\caption{Efficient variance selection}
\label{alg:eff_var}
\end{algorithm}
%We use the values 2 and 1000 for $\tau_{BF}$ and $\tau_{FB}$ respectively. The
Algorithm 1 for efficient computation results in a reduction in computation in
about 80\% of the pixels in the videos when $\tau_{BF}$ is set to 2, with a
slight reduction in the F-measure by about 1 to 2\% on most videos when compared
to the full implementation. The efficient variance selection procedure however
still performs significantly better than the uniform variance model by 2 to 10\%
on most videos. 
%Our unoptimized matlab code (without caching) takes $10$ seconds per frame for a
%128x160 sized video. Uniform kernel variance (our implementation of Sheikh and Shah) takes 5 seconds. The caching algorithm takes about $6$ seconds, making it very comparable to Sheikh and Shah. 

%Since the MRF cleaning procedure is another
%computationally complex operation, we report results from turning off the MRF
%cleaning procedure and using the above caching scheme in the last column of
%The resulting accuracy from using this caching scheme is shown in
%column 9 of table \ref{tbl:I2R_compare}. Although there is a reduction in
%accuracy compared to DF(rgb), the results are still more accurate than MoG and
%KDE. Also, the algorithm continues to be more accurate than SILTP on the videos
%with large objects.
\section{Discussion}\label{sec:disc}
By applying kernel estimate method to a large data set, we have established, as
do Sheikh and Shah~\cite{Sheikh05}, that the use of spatial information is extremely helpful. 
Some of the important issues pertaining to the choice of kernel parameters for
data sets with wide variations have been addressed. Having a uniform
kernel variance for the entire data set and for all pixels in the image results
in a poor overall system. Dynamically adapting the variance for each pixel 
results in a significant increase in accuracy. 

%Inspite of using basic color intensities as features,
%\DF performance is comparable to state-of-art features that are designed to
%handle shadows and lighting changes. 

Using color features in the joint domain-range kernel estimation approach can
complement complex background model features in settings where the
latter are known to be inaccurate. Combining robust color features like LAB with
texture features like SILTP in a VKS framework yields a highly accurate
background classification system.

%For future work, we believe tracking of foreground objects and incorporating
%tracking information in the foreground scores and prior estimates 
%would improve the system further.
For future work, we believe our method could be explained more elegantly in a 
probabilistic framework where the scores are replaced by likelihoods and 
informative priors are used in the Bayes rule classification.
%Equation \ref{eq:bayes_score_e} provides sound classification decisions in the
%absence of supporting evidence for either class. It could be useful in other
%classification problems to bias the decision in favor of a particular class
%when the likelihoods are inconclusive.
%Another advantage of using a VKS approach to backgrounding is that it lends
%itself to application to the tracking problem easily. By defining a \DF for
%objects and using informative priors and a search method to locate objects, the
%background models and tracking models can be integrated in a combined
%framework.

%Using support from spatially nearby pixels in classification could lead to a
%background model that is more robust to jitter.  By using a spatial variance
%equal to the amount of expected jitter in the video, we expect to model jitter
%in videos.
\section{Acknowledgements}\label{sec:ack}
This work was supported in part by the National Science Foundation under CAREER award
IIS-0546666 and grant CNS-0619337. Any opinions, findings, conclusions, or recommendations expressed here are the author(s) and do not necessarily reflect those of the sponsors.

{\small
\bibliographystyle{ieee}
\bibliography{bib_df_kde_bg_final}
}

\end{document}